# Feature Importance in Bayesian Assessment of Newborn Brain Maturity from EEG

L. JAKAITE, V. SCHETININ, and C. MAPLE
Department of Computer Science and Technology
University of Bedfordshire
Luton, LU1 3JU
UK
Vitaly.Schetinin@beds.ac.uk

*Abstract:* - The methodology of Bayesian Model Averaging (BMA) is applied for assessment of newborn brain maturity from sleep EEG. In theory this methodology provides the most accurate assessments of uncertainty in decisions. However, the existing BMA techniques have been shown providing biased assessments in the absence of some prior information enabling to explore model parameter space in details within a reasonable time. The lack in details leads to disproportional sampling from the posterior distribution. In case of the EEG assessment of brain maturity, BMA results can be biased because of the absence of information about EEG feature importance. In this paper we explore how the posterior information about EEG features can be used in order to reduce a negative impact of disproportional sampling on BMA performance. We use EEG data recorded from sleeping newborns to test the efficiency of the proposed BMA technique.

## 1 Introduction

Assessment of brain maturity can be obtained by estimating newborn's age from sleep EEG [1] - [3]. This approach is based on the clinical evidences that the post-conceptional and EEG estimated ages of healthy newborns typically match each other, and the newborn's brain maturity is most likely abnormal if the ages mismatch [2], [4]. Thus, the mismatch alerts about abnormal brain development.

The established assessment methodologies are based on learning models from EEGs recorded from sleeping newborns whose brain maturity was already assessed by clinicians. The regression models are made capable of mapping the brain maturity into EEG based index [5]. The classification models are made capable of distinguishing maturity levels: at least one with normal and other with abnormal brain maturity [4], [6]. The established methodologies are based on learning a single model from a given set of data and they cannot ensure that a model will not be overfitted to the data

Probabilistic reasoning, based on the Bayesian methodology of averaging over decision models, enables to evaluate the uncertainty in decision making [7] – [9]. The use of the Bayesian Model Averaging (BMA) over Decision Trees (DTs) enables to select features which make the most significant contribution to outcomes, and the resultant DT ensemble can be interpreted by clinicians as shown in [10]. However, the success in implementation of BMA is critically dependent on the diversity and proportion of models sampled for averaging. The models should be diverse in parameters and structure. The portion of models whose likelihood is high should be largest to ensure unbiased estimates. The use of a priori information provides better conditions for achieving these requirements. In many practical cases when a priori information about feature importance is absent, the provision of the diversity and proportion of models becomes problematic [11], [12].

In our previous research [13], we attempted to overcome the above BMA problems and developed a new feature selection strategy for Bayesian averaging over DTs to assess trauma patient survival. In emergency care, trauma patient's condition should be urgently evaluated by a clinician through a screening procedure which typically comprises around 20 tests. If the screening tests are ambiguously interpreted, and evaluation of the severity of the injury is misleading, the mistake in a decision can be fatal for the patient. We supposed that some tests make a weak contribution to the outcome, and therefore avoidance of such tests can improve the assessment accuracy. In contrary, we observed that the performance had decreased and then assumed that this could happen because the discarded weakest attribute was still important for some portion of the data. An alternative assumption was that the weakest attribute makes a noticeable contribution to the learning results. It is important for clinical practice to reduce the number of screening tests required for making reliable decisions. In our experiments we used the UK Trauma data to test the efficiency of the proposed strategy and then found that it enables to reduce the number of screening tests, keeping the performance and reliability of making decisions high. In this paper we aim to further explore this strategy being applied to the problem of EEG assessment of newborn brain maturity.

The paper is structured as follows. Section 2 states the problem in Bayesian assessment of newborn EEG maturity. Section 3 describes the bases of BMA over DTs, and section 4 describes the EEG data used for the experiments. The experimental results are presented in section 5, and section 6 concludes the paper.

## 2 Problem Statement

The newborn brain maturity can be estimated by experts from sleep EEG in terms of post-conceptional age measured in weeks. The accuracy of such assessment is typically two weeks. We aim to develop a BMA technique to assess the brain maturity from sleep EEGs which are typically recorded via the standard C3-C4 electrode system. These recordings were transformed into the standard frequency bands listed in Table 1. The spectral features along with their statistical characteristics form a multidimensional representation of the EEG data. The spectral features are presented by the absolute and relative spectral powers calculated for electrodes C3, C4, and their sum. Additionally the statistical variances are calculated for these features, so that the total number of EEG attributes becomes 72. No other information about the data is available to make assumptions on EEG feature importance.

Under these conditions the standard BMA techniques being used for assessment of the EEG maturity cannot ensure unbiased results. This mainly happens because the detailed exploration of a multidimensional model space becomes problematic within a reasonable time. The detailed exploration is needed to ensure that the majority of models are sampled from areas of maximal posterior. Otherwise, instead of exploring all possible areas of maximal posterior, the models will be disproportionally sampled from some of these areas. The negative effect of disproportional sampling of models is that results of BMA become biased.

Obviously, information about feature importance can reduce a model parameter space that needs to be explored. However in our case, we have no such information, and therefore we are forced to make an unrealistic assumption that all the EEG attributes make an equal contribution the age assessment. Fortunately, DT models provide the feature selection, and the use of Bayesian averaging over such models gives a posterior information about feature importance. This means that if a feature is rarely used in the models, then we conclude that this attribute makes a weak contribution and should be deleted.

Clearly, when there are few weak attributes, the portion of models using such attributes is small, and their impact on the outcome is negligible. On the contrary, when the number of weak attributes is large, the disproportion in models becomes significant. Therefore we could improve the BMA results by reducing the disproportional sampling. In this research we aim to explore whether discarding the models using weak EEG attributes will reduce the bias in the assessment of brain maturity.

A trivial strategy of using the posterior information for feature selection within BMA is to use this information to learn a new ensemble from a data set in which the weak attributes were deleted. This strategy reduces a model parameter space, and therefore it enables to explore this space in more detail. The other strategy that can be thought of is refining the ensemble by discarding models which use weak attributes. We expect that such refinement can improve the BMA performance.

## 3 Implementation

For a DT given with parameters $\theta$, the predictive distribution is written as an integral over the parameters $\theta$.

$$p(y\,|\,\mathbf{x},\mathbf{D}) = \int_\theta p(y\,|\,\mathbf{x},\theta,\mathbf{D}) p(\theta\,|\,\mathbf{D}) d\theta,$$

where $y$ is the predicted class (1, ..., $C$), $x = (x_1, ..., x_m)$ is the m-dimensional vector of input, and $\mathbf{D}$ are the given training data.

This integral can be analytically calculated only in simple cases, and in practice part of the integrand, which is the posterior density of $\theta$ conditioned on the data $\mathbf{D}$, $p(\theta\,|\,\mathbf{D})$, cannot usually be evaluated. However, for $\theta(1), ..., \theta(N)$ are the samples drawn from the posterior distribution $p(\theta\,|\,\mathbf{D})$, we can write.

$$p(y \mid \mathbf{x}, \mathbf{D}) \approx \sum_{i=1}^{N} p(y \mid \mathbf{x}, \theta^i \mathbf{D}) p(\theta^i \mid \mathbf{D}) = \frac{1}{N} \sum_{i=1}^{N} p(y \mid \mathbf{x}, \theta^i, \mathbf{D}).$$

The above integral can be approximated by using Markov Chain Monte Carlo (MCMC) technique as described in [7], [9]. To perform such an approximation, we need to run a Markov Chain until it has converged to a stationary distribution. Then we can collect N random samples from the posterior $p(\theta \mid \mathbf{D})$ to calculate the desired predictive posterior density.

Using DTs for the classification, we need to find the probability φtj with which an input x is assigned by terminal node $t = 1, \ldots, k$ to the $j$th class, where $k$ is the number of terminal nodes in the DT. The DT parameters are defined by $s_i^{pos}$, $s_i^{var}$, $s_i^{rule}$, $i = 1, \ldots, k - 1$, where $s_i^{pos}$, $s_i^{var}$, and $s_i^{rule}$ define the position, predictor and rule of each splitting node, respectively. For these parameters the priors can be specified as follows. First, we can define a maximal number of splitting nodes, $s_{max} = n - 1$. Second we draw any of the $m$ attributes from a uniform discrete distribution $U(1, \ldots, m)$ and assign $s_i^{var} \in \{1, \ldots, m\}$.

Finally the candidate value for the splitting variable $x_j = s_i^{var}$ can be drawn from a discrete distribution $U(x_j(1), \ldots, x_j(L))$, where $L$ is the number of possible splitting rules for variable $x_j$. Such priors allow us to explore DTs which split data in as many ways as possible. However the DTs with different numbers of splitting nodes should be explored in the same proportions [7], [9].

To sample DTs of a variable dimensionality, the MCMC technique exploits the Reversible Jump extension. To implement the RJ MCMC technique, in [7], [9] it has been suggested exploring the posterior probability by using the following types of moves:

Birth. Randomly split the data points falling in one of the terminal nodes by a new splitting node with the variable and rule drawn from the corresponding priors.

Death. Randomly pick a splitting node with two terminal nodes and assign it to be one terminal with the united data points.

Change-split. Randomly pick a splitting node and assign it a new splitting variable and rule drawn from the corresponding priors.

Change-rule. Randomly pick a splitting node and assign it a new rule drawn from a given prior.

The first two moves, birth and death, are reversible and change the dimensionality of θ. The remaining moves provide jumps within the current dimensionality of θ. Note that the change-split move is included to make "large" jumps which potentially increase the chance of sampling from a maximal posterior whilst the change-rule move does "local" jumps.

The RJ MCMC technique starts drawing samples from a DT consisting of one splitting node whose parameters were randomly assigned within the predefined priors. So we need to run the Markov Chain while a DT grows and its likelihood is unstable. This phase is said *burn-in* and it should be preset sufficiently long in order to stabilize the Markov Chain. When the Markov Chain will be enough stable, we can start sampling. This phase is said *post burn-in*.

## 3 The Proposed Method

To test the assumption made in section 2 and refine DT model ensembles obtained with BMA, we propose a new strategy aiming at discarding the DT models which use weak attributes. According to this strategy, first the BMA technique described in section 2 is used to collect DT models. Then posterior probabilities of using attributes in the ensemble of DT models are estimated. These estimates give us the posterior information on feature importance. Having obtained a range of the posterior probabilities, we then define a threshold value to cut off the attributes with the probabilities below this threshold – we define such attributes as weak. At the next stage we find the DT models which use these weak attributes and finally discard these DT models from the ensemble.

Obviously, the larger the threshold value, the greater number of attributes is defined as weak, and therefore the larger portion of DT models is discarded. The efficiency of this discarding technique is evaluated in terms of the accuracy of the refined DT ensemble on the test data. The uncertainty in the ensemble outcomes is evaluated in terms of entropy. Having a set of the threshold probability values obtained in a series of experiments, we can expect that there is an optimal threshold value at which the performance becomes higher. We can also expect to find a threshold value at which the uncertainty becomes lower. In the following section we test the proposed technique on the problem of assessment of newborn brain maturity from sleep EEG.

## 4 Experiments

In our experiments we used EEG data recorded from 686 newborns during sleep hours. The newborns were aged from 40 to 45 weeks post-conception. Each of these 6 groups contains around 100 patients. The EEGs have been segmented in 10-s intervals to be represented by 72 attributes as spectral powers and their variances. We averaged the EEG segments of each patient to represent the patient by one data sample, so that the problem was represented by 686 data samples in 72 attribute space.

For experiments we used the Bayesian averaging over DT models introduced in section 3 and described in our previous publications [12]. The BMA ran with the following settings. In a burn-in phase we collected 200,000 DTs, and in a post burn-in phase 10,000 DTs. During the post burn-in phase each 7th model was collected to reduce the correlation between DT models. The minimal number of data samples allowed in DT nodes was 6. Proposal variance was 1.0, and probabilities of making moves of birth, death, change variable, and change threshold were set to 0.15, 0.15, 0.1, and 0.6, respectively. The performance and uncertainty of the DT ensemble collected in the post burn-in phase were evaluated within a 3-fold cross-validation and $\pm 2\sigma$ intervals.

The BMA with the above settings has accepted around 0.13 of DT models in both phases. The DT size becomes stationary around 30 nodes soon after 10,000 samples of burn-in phase. The average performance was 27.4%, whilst the accuracy of classification by chance was $100/6 = 16.7\%$. The performance varied within $2\sigma$ interval equal to 8.2%. The entropy of the DT ensemble was $478.3 \pm 15.8$.

According to the proposed technique, we estimated the importance of all the 72 attributes in terms of the posterior probabilities of using these attributes by the DT models collected in the post burn-in phase. The posterior probabilities of using the attributes ranged between 0.0 and 0.005. We then gradually increased the threshold value $T$ from 0 at steps of 0.001 to 0.005 to define features as weak accordingly to the proposed strategy of feature selection. From Table 1, we can see that at threshold value 0.001 the average number of weak attributes, $k$, was 14, whilst at level 0.005 their number has increased to 31.

Having found the weak attributes, we applied the proposed technique to refine the DT ensemble. Table 1 shows the number of weak attributes, $k$, versus the threshold values, $T$, within a 3-fold cross-validation. From Table 1 we can see that the performance $P$ of the refined DT ensemble is slightly increased from 27.4 to 29.2 when the threshold is gradually increased from 0.0 to 0.005. At the same time the uncertainty in decisions is decreased from 478.4 to 469.0 in terms of entropy $E$ of the ensemble. For comparison, we applied a technique of discarding the same weak attributes and then reran the BMA on the data reduced in their dimensionality.

From Table 1 we can see that the BMA performance has slightly increased from 27.4 to 29.0 when 23 weak attributes were discarded. The discarding of 31 attribute has resulted in a decrease in the ensemble entropy from 478.3 to 463.6. Overall, the both techniques are shown to provide the comparable performance and ensemble entropy. However, the technique of discarding attributes has shown to tend to perform in a larger variation. Within this technique for each threshold value it is required to retrain DT ensemble on the data of a new dimensionality.

**Table 1.** Performance (P) and entropy (E) of the techniques versus threshold values (T) within 3-fold cross-validation.

| $T$ | $k$ | Proposed technique | | Technique of discarding attributes | |
|---|---|---|---|---|---|
| | | $P$, % | $E$ | $P$, % | $E$ |
| 0.001 | 14 | 27.5±8.4 | 478.4±15.8 | 28.7±9.6 | 469.0±13.7 |
| 0.002 | 18 | 27.8±9.0 | 477.7±16.4 | 25.8±1.7 | 475.7±16.7 |
| 0.003 | 23 | 28.7±8.2 | 475.7±15.3 | 29.0±8.5 | 474.1±33.9 |
| 0.004 | 28 | 28.9±7.6 | 471.2±10.3 | 28.4±1.8 | 472.4±12.0 |
| 0.005 | 31 | 29.2±7.9 | 469.0±11.9 | 27.3±6.5 | 463.6±26.3 |

Fig 1 shows the distributions of performances calculated for the original and refined DT ensembles. According to the proposed method, the refinement has been obtained by discarding those DTs which use 26 weak attributes. We can see that the size of the refined ensemble becomes significantly smaller. Most of the DTs with performance above 32.0% have been kept, whilst most of the DTs with performance below 24.0% have been discarded from the refined ensemble.

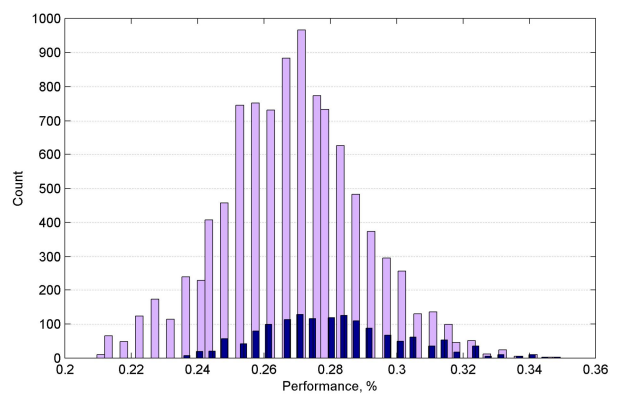

**Fig. 1.** Distributions of performances of DTs included in the original (in grey) and refined (in black) ensembles.

## 4 Conclusion and Discussion

Under the lack of prior information, BMA MCMC technique tends to sample from the posterior disproportionally that affects the BMA performance. In this paper we explored how the posterior information about attributes can be employed in order to reduce a negative impact of disproportional sampling on BMA performance. We assumed that the posterior information about feature importance can be used to find weak attributes, and we proposed a new technique aiming at refining an ensemble by discarding the models which use the weak attributes.

According to our assumption, in the presence of weak attributes some models included in the resultant ensemble will use these weak attributes. The larger the number of weak attributes, the greater the negative impact on BMA performance. We expect that the discarding of models using weak attributes will reduce the bias and improve the BMA performance.

To test the proposed technique we used EEG data recorded from sleeping newborns. Our experiments have shown that the proposed technique is capable of increasing the BMA performance and decreasing the ensemble entropy. We observed that the proposed technique enables DTs with higher performance to be included in the ensemble while discarding the DTs with lower performance. Thus the proportion of DT models included in the ensemble is improved due to decreasing the number of DTs with lower performance. We observed that the MCMC technique makes a candidate model acceptable with different attributes. An accepted model may include by chance a weak attribute even with a small decrease in performance. In the presence of many weak attributes chances of accepting a model which includes a weak attribute are increased, and this leads to a disproportion of models in the ensemble.

Typically, a technique of reduction of the data dimensionality by discarding of the weak attributes is expected to improve BMA performance due to reducing a model parameter space needed to be explored. However this technique requires rerunning BMA. The proposed technique was shown to provide the comparable performance without the need of rerunning the BMA.